\documentclass[pdflatex,sn-mathphys-num]{sn-jnl}

\usepackage{amsmath,amssymb,amsfonts}%
\usepackage{mathtools}
\usepackage{caption}
\usepackage{subcaption}
\usepackage{graphicx}%
\usepackage{multirow}%
\usepackage{amsthm}%
\usepackage{mathrsfs}%
\usepackage[title]{appendix}%
\usepackage{xcolor}%
\usepackage{textcomp}%
\usepackage{manyfoot}%
\usepackage{booktabs}%
\usepackage{algorithm}%
\usepackage{algorithmicx}%
\usepackage{algpseudocode}%
\usepackage{listings}%

\theoremstyle{thmstyleone}%
\theoremstyle{thmstyletwo}%

\theoremstyle{thmstylethree}%

\begin{document}

\title[Exploring the Relationship Between Feature Attribution Methods]{Exploring the Relationship Between Feature Attribution Methods and Model Performance}

\author*[1,3]{\fnm{Priscylla} \sur{Silva}}\email{priscylla.silva@usp.br}

\author[2]{\fnm{Claudio} \sur{Silva}}\email{csilva@nyu.edu}

\author[1]{\fnm{Luis Gustavo} \sur{Nonato}}\email{gnonato@icmc.usp.br}

\affil*[1]{\orgname{University of São Paulo}, \orgaddress{\city{São Carlos}, \state{SP}, \country{Brazil}}}

\affil[2]{\orgname{New York University}, \orgaddress{\city{Brooklyn}, \state{NY}, \country{USA}}}

\affil[3]{\orgname{Federal Institute of Alagoas}, \orgaddress{\city{Rio Largo}, \state{AL}, \country{Brazil}}}

\abstract{Machine learning and deep learning models are pivotal in educational contexts, particularly in predicting student success. Despite their widespread application, a significant gap persists in comprehending the factors influencing these models' predictions, especially in explainability within education. This work addresses this gap by employing nine distinct explanation methods and conducting a comprehensive analysis to explore the correlation between the agreement among these methods in generating explanations and the predictive model's performance. Applying Spearman's correlation, our findings reveal a very strong correlation between the model's performance and the agreement level observed among the explanation methods.}


\maketitle

\section{Introduction}
\label{sec:intro}
Extensive research has been conducted on applying machine and deep learning methods in education. These methods encompass a wide range of automated processes, from grading assignments to generating tailored feedback~\citep{SUZEN2020726, BERNIUS2022100081}. Predicting student success and early course dropout is particularly crucial~\citep{su14106199}. Models that address these issues try to identify which students are at higher risk of failing or dropping out~\citep{data7110146, NIYOGISUBIZO2022100066, Mubarak2022}. By utilizing these models, educators can proactively intervene and provide tailored support to help students succeed in their coursework. These models analyze student information, such as academic records, engagement, and demographic data, to find patterns predicting future academic outcomes.

Considerable effort has gone into refining the accuracy of predictive models. However, there remains a knowledge gap regarding the inner workings of these models. It is insufficient to identify a potential student failure only; it is necessary to identify the factors analyzed by the model to generate the predictions.~\citet{9720806} present a predictive model for student success in secondary education using various classification algorithms; the study emphasizes the importance of interpretability and transparency in model predictions, employing LIME (Local Interpretable Model-agnostic Explanations) to enhance understanding of the model predictions.~\citet{Baranyi2020} conducted a study that aimed to predict the risk of college students dropping out at the Budapest University of Technology and Economics. They employed advanced machine learning models, including deep neural networks and gradient-boosted trees, and focused on interpreting the models by using two techniques - permutation importance and SHAP values. The study sheds light on the importance of model interpretation in predicting student dropout risk.

Predicting student success is challenging, and many models used for this purpose are difficult to interpret because of their black-box nature. This lack of transparency makes it hard to understand how decisions are made and what factors contribute to making predictions. As a result, it is not easy to gain meaningful insights into the factors that impact student success. However, the explainable machine learning community has made significant progress in developing different methods to elucidate the inner workings of models. Some of these methods focus on local explanation techniques that delve into the intricacies of model predictions at an individual instance level. One prominent avenue within local explanation methods involves elucidating feature importance. By employing these techniques, practitioners can gain insights into the importance of each input feature in influencing model predictions. Significant efforts have been made to use explanation methods to understand how a model predicts student success. However, according to~\citet{2022_EDM_long_papers}, there is a considerable gap in the literature when it comes to explaining the results in the field of education.

\citet{krishna2022disagreement} have highlighted a significant concern associated with feature attribution explanation methods known as the \textit{disagreement problem}. This issue arises from the notable disparity in identifying the most important features among various explanation methods. The critical nature of the \textit{disagreement problem} becomes evident when considering the implications: if distinct methods yield divergent explanations, the question of trustworthiness arises. In the context of education, specifically within student success prediction,~\citet{2022_EDM_long_papers} present compelling evidence that different explanation methods applied to the same model and course yield markedly distinct feature importance distributions. This underscores the gravity of the \textit{disagreement problem} in educational scenarios, raising crucial questions about the reliability and consistency of explanatory insights derived from these methods. The \textit{disagreement problem} remains unresolved in the existing literature. Our primary goal is to address the following research question: Is there a correlation between the model's performance and the disagreement level observed among explanation methods? To achieve this, we used nine popular instance-based explanation techniques to predict student success in two distinct real-world datasets.

\section{Methodology}
\label{sec:methodology}
In this section, we will define the task of predicting student success, the \textit{disagreement problem} that arises when using different explanation methods, and the metrics used to measure the (dis)agreement level between these methods. We will then introduce this study's datasets, model training, and explanation methods. Finally, we will describe our experiment setup in detail.

\subsection{Problem Formulation}

In this study, we are considering the student success prediction as a binary classification task. Let $X$ be the feature space, representing the input features of a student. The feature vector for a particular student is denoted as $x \in X$. Let $Y$ be the label space, where $y \in \{0, 1\}$ represents the binary outcome of student success. Here, $y = 1$ may signify success, while $y = 0$ denotes otherwise. A binary classification model is a function $f: X \rightarrow [0, 1]$ that assigns a probability to each instance, indicating the likelihood of success.

A local attribution method for the model $f$ is a mapping $g: (f,X) {\rightarrow} E$ that, based on $f$, takes instances from $X$ to the explanation space $E$, where $g(x)=(e_1,\ldots,e_K)$ is a point in $E$, $K$ denotes the number of features, and $e_i$ are the importance of each feature as to $f$. Consider two distinct local attribution methods, $g_1$ and $g_2$. For a given instance $x$, let $g_1(x)$ and $g_2(x)$ be the explanations generated by $g_1$ and $g_2$ in $x$, respectively. The disagreement problem occurs when $g_1(x) \neq g_2(x)$.

\citet{krishna2022disagreement} introduced a set of metrics to measure the (dis)agreement between two local attribution explanations. The metrics evaluate (dis)agreement in the top-$k$ most important features identified by two explanation methods. Our focus in this study is on the metrics, namely, Feature Agreement (FA), Sign Agreement (SA), Rank Agreement (RA), and Signed Rank Agreement (SRA).

Feature Agreement (FA) determines the proportion of common features between the sets of top-$k$ features in two explanations. Sign Agreement (SA) assesses the proportion of common features with the same sign among the top-$k$ features of two explanations.
The positive and negative signs indicate the effect of a feature on the model's prediction. A positive attribution score means a feature contributes positively, while a negative score indicates the opposite.
Rank Agreement (RA) calculates the fraction of common features in the same position of the rank of importance among the top-$k$ features of two explanations. Signed Rank Agreement (SRA) combines the previous methods, incorporating both rank and sign. All the metrics listed above are in the interval [0, 1], with zero indicating complete disagreement and one representing total agreement. Additional information on the metrics can be found in Appendix~\ref{apd:disagreement_metrics}.

\subsection{Experimental Setup}
In our experiment, we utilized two datasets. The first dataset was provided by~\citet{Amrieh_2015} and consisted of 480 students and 16 predictive features collected from a Kalboard 360 e-learning system. The target of this dataset is a multiclass label that classified student grades into low, medium, and high categories. After the data preprocessing step (see Appendix~\ref{apd:datasets} for more details on preprocessing), we were left with 12 features. Since we were working with binary classification, we only used students classified in the low and high categories, where high represented the positive class, leaving us with a dataset of 269 students. The second dataset was collected from a group of 132 computer science and computer engineering students taking an Introduction to Programming course during their first semester at a university in Brazil. It consists of 16 predictive features, and the target label is binary, indicating whether the student passed or failed the course (see Table~\ref{tab:dataset2} in Appendix~\ref{apd:datasets} for more details on features).

We trained Neural Network models for each dataset. The model for the~\citet{Amrieh_2015} dataset consists of two hidden layers, with 16 and 8 neurons, respectively. On the other hand, the model for the Introduction to Programming course dataset includes two hidden layers with 32 and 16 neurons, respectively.  In the experiment, we used 70\% of the data for training, 15\% for validation, and 15\% for testing. Throughout the training, we systematically saved the models from intermediate epochs, creating a series of snapshots that captured the evolving state of the neural network.

We employed nine state-of-the-art feature attribution techniques to explain the predictions made by the models for the data in the testing set. These methods included six gradient-based methods, namely DeepLift~\citep{deeplift}, Guided Backprop~\citep{guidedback}, Input X Gradient~\citep{inputxgrad}, Integrated Gradients~\citep{integratedgrad}, Smooth Gradient~\citep{smilkov2017smoothgrad}, and Vanilla Gradients~\citep{vanillagrad}, along with three other techniques named LIME~\citep{lime}, Occlusion~\citep{occlusion}, and KernelShap~\citep{kernelshap}. For further details, see the Appendix~\ref{apd:explanationmethods}.

\subsubsection{Model Performance in Intermediate Epochs}

Using the test data, we computed the Area Under the Receiver Operating Characteristic Curve (AUC) metric for each model from the intermediate epochs. The AUC metric is a valuable measure for binary classification models, quantifying the model's ability to distinguish between positive and negative instances across different probability thresholds. Calculating the AUC at each intermediate epoch, we obtained a dynamic model performance profile throughout training.

\subsubsection{(Dis)agreement measurement}

We used the saved models to predict the test set data at each intermediate epoch. Then, we applied the selected explanation methods to generate explanations for each individual prediction. These methods generated importance scores for the features of each instance in the test set, offering a detailed understanding of the contribution of each input feature to the model's decision-making process. 

We employed established (dis)agreement metrics (FA, SA, RA, and SRA) to systematically quantify the (dis)agreement level between the explanation methods. These metrics operate on a per-instance basis, so we calculated the average (dis)agreement across all instances in the test set, providing a comprehensive assessment of the overall agreement among the selected explanation methods. This process was repeated for every model stemming from the intermediate epochs, enabling us to discern patterns in the evolution of explanation methods disagreements throughout the neural network's training. By averaging the (dis)agreement scores for all instances, we obtained a robust measure of the consensus or divergence among the explanation methods. Appendix~\ref{apd:heatmap} provides an example of the disagreement between the pairs of methods for the two datasets used in our study. Additionally, Appendix~\ref{apd:boxplot} provides an example of how the distribution of the (dis)agreement scores can vary.

\subsubsection{Correlation Analysis}

In order to evaluate the (dis)agreement metrics accurately, it is necessary to vary the value of $k$ between 1 and the total number of features present in each dataset. This is because the size of the top-$k$ features significantly impacts the (dis)agreement metrics.

We compute Spearman's rank correlation to explore the relationship between model performance, as measured by the AUC metric, and the (dis)agreement level. In Figure~\ref{fig:corr_cs101}, the results are presented for the Introduction to Programming course dataset, with columns representing the four metrics used to measure (dis)agreement and lines showcasing the variation in $k$ used in the top-$k$. Each dot on the charts corresponds to models from the intermediate epochs.

\section{Results and Discussions}

In this section, we analyze the relationship between model performance and (dis)agreement level among the employed explanation methods by examining the Spearman correlation results. A higher AUC value indicates better model performance, while a higher disagreement metric value indicates stronger consensus among methods.

\begin{figure}[ht]
\centering
\includegraphics[width=1\textwidth]{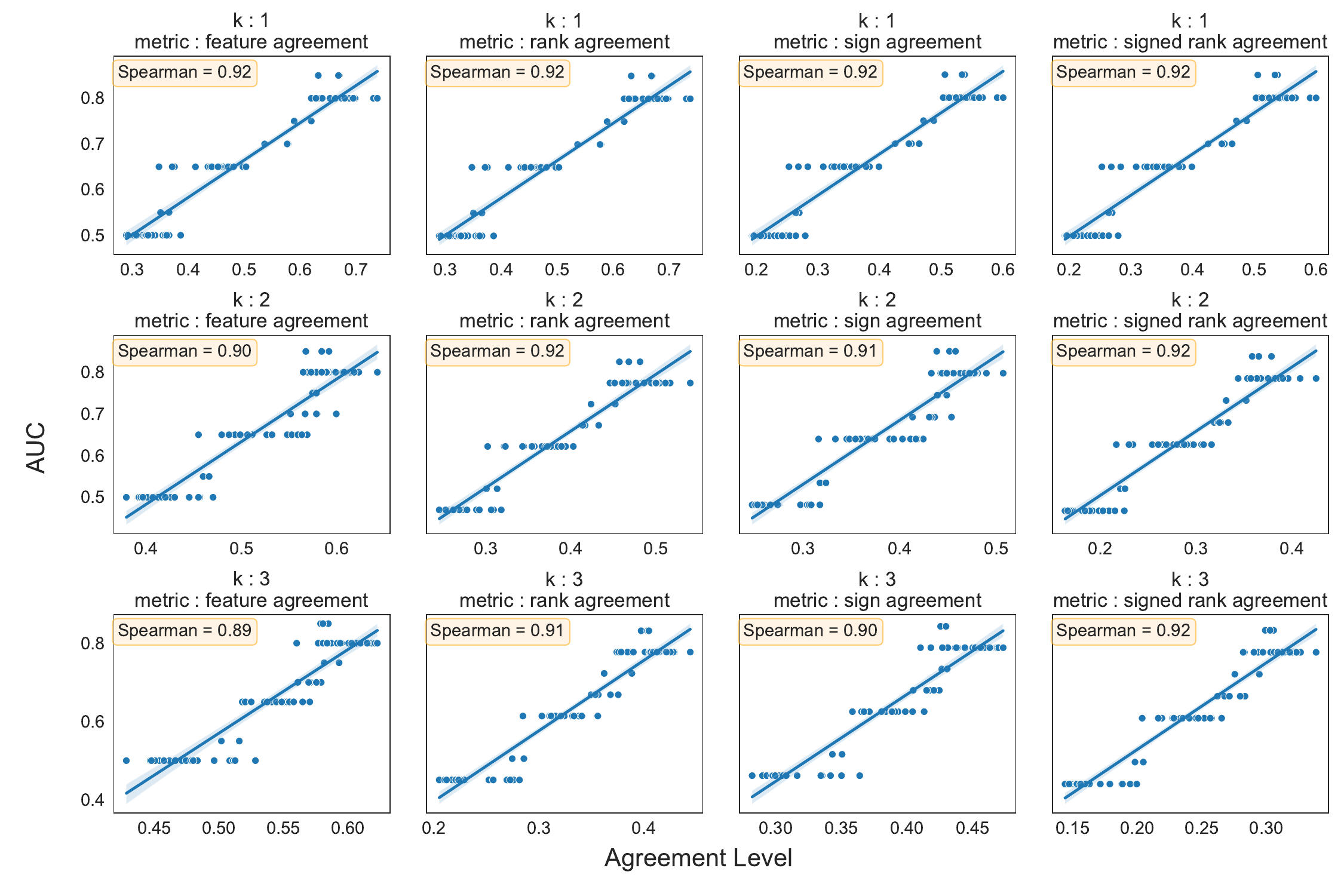}
\caption{Correlation between Model Performance (AUC) and (Dis)agreement Metrics for Models Trained on the Introductory Programming Course Dataset.}\label{fig:corr_cs101}
\end{figure}

Our findings are visually represented in Figures~\ref{fig:corr_cs101} and~\ref{fig:corr_xapi}, with the x-axis denoting the (dis)agreement level and the y-axis representing the model's performance. The charts are organized as follows: columns present charts for each (dis)agreement metric, while rows showcase the variation in the $k$ values used for calculating the metric.

Figure~\ref{fig:corr_cs101} shows a sample of the results for models trained on the dataset of the n Introduction to Programming course. In the figure, $k$ ranges from 1 to 3\footnote{For the complete figure with $k$ ranging from 1 to the total number of features, please refer to Appendix~\ref{apd:corr}}. In 87.5\% of cases for this dataset, the Spearman correlation values surpassed 0.8, indicating a robust and consistent correlation. Correlations below 0.8 were primarily observed for the FA metric as the $k$ value increased, approaching the total number of features in the dataset. The unique behavior of the FA metric explains this phenomenon. Specifically, when $k$ equals the number of features, the FA metric results in 100\% agreement between explanations. This characteristic arises from the metric considering the intersection between sets of top-$k$ features from two explanation methods. When $k$ aligns with the total number of features, the two sets become identical, yielding unanimous agreement between explanations.

Figure~\ref{fig:corr_xapi} presents a sample of the study outcomes on~\citet{Amrieh_2015}'s dataset, where $k$ ranges from 10 to 12. Our analysis revealed that for this dataset, in 79\% of cases, the Spearman correlation score surpassed 0.8, indicating a very strong positive correlation between AUC and the level of agreement. In 16\% of cases, the Spearman correlation fell between 0.5 and 0.8, signifying a strong correlation. As we previously discussed, the correlation dropped below 0.5 in certain cases related to the FA metric, which was consistent with the behavior observed before. It has been noted before that when $k$ equals the number of features, the (dis)agreement level is always 1, regardless of the model's performance (as shown in the chart in the first column and third line). This highlights the behavior of the FA metric. Appendix~\ref{apd:corr} contains the complete figure.

\section{Discussion and Conclusion}

In the results section, we showed that for both datasets analyzed in the student success prediction task, we were able to observe that there is a strong correlation between the model's performance, measured using AUC, and the (dis)agreement level between the methods, measured using the FA, SA, RA, and SRA metrics. The strong correlation we identified implies that the agreement among explanation methods becomes more evident as the model's performance improves. A higher-performing model tends to yield explanations that exhibit more substantial consensus across various explanation techniques. This finding underscores the intrinsic connection between model quality and the interpretability of its predictions.

\begin{figure}[ht]
\centering
\includegraphics[width=1\textwidth]{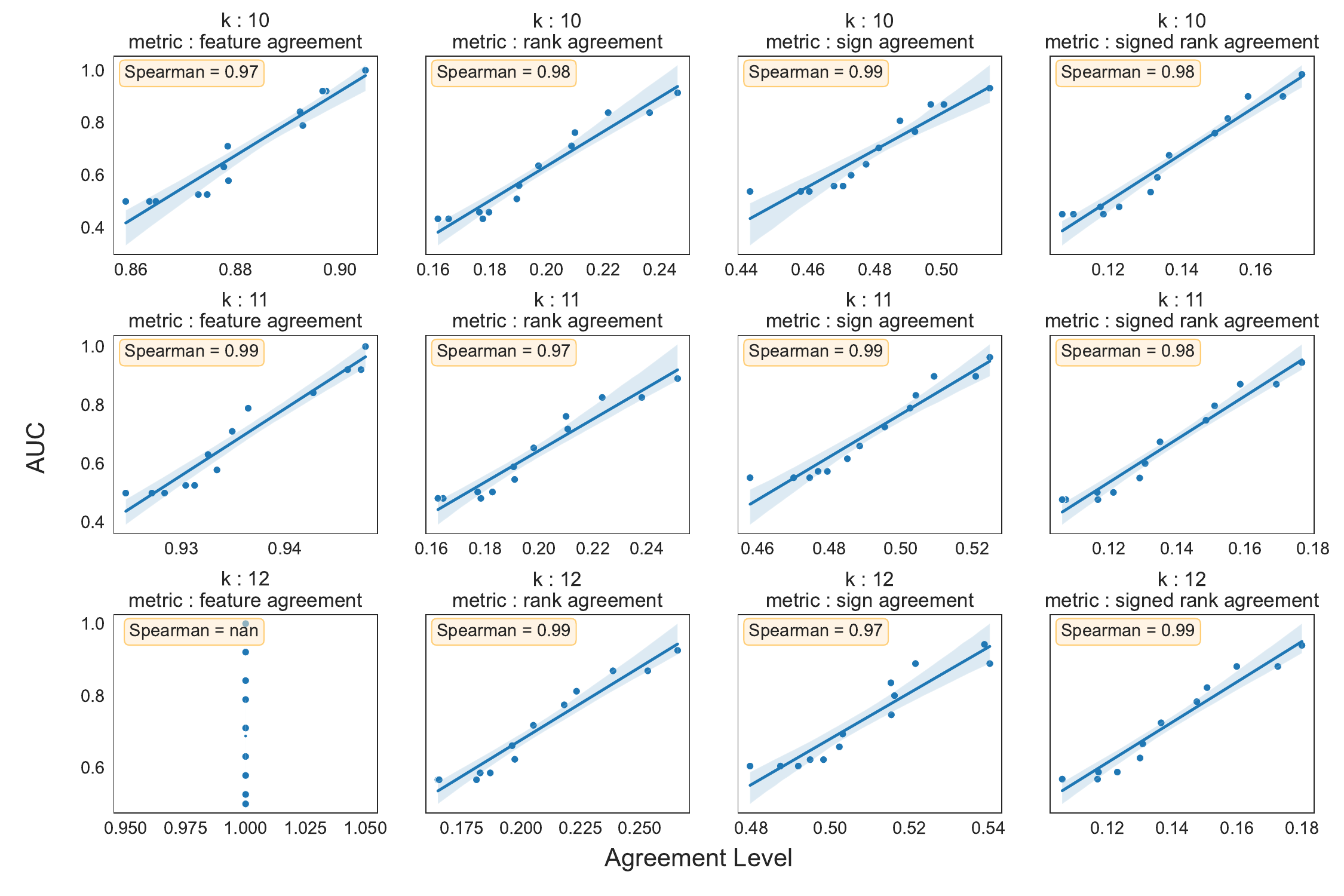}
\caption{Correlation between Model Performance (AUC) and (Dis)agreement Metrics for Models Trained on~\citet{Amrieh_2015}'s Dataset.}\label{fig:corr_xapi}
\end{figure}

Our results have significant implications for practitioners and experts using explanation methods. Notably, we advocate for thoughtful consideration of the model's performance before employing any explanation method. Figures~\ref{fig:corr_cs101} and~\ref{fig:corr_xapi} depict this relationship, illustrating that models with an AUC greater than or equal to 0.8 consistently exhibit the highest levels of agreement among explanation methods. In conclusion, our study emphasizes the intertwined nature of model performance and explainability, reinforcing the importance of a robust model before delving into the realm of explanation methods. By prioritizing models with AUC values above the 0.8 threshold, practitioners can enhance the reliability and coherence of explanations generated by various methods.

\backmatter

\bmhead{Acknowledgements}

This work was supported by FAPESP grants  2022/09091-8, 2023/05783-5, 2022/03941-0 and CNPq grant 307184/2021-8. The opinions, hypotheses, and conclusions or recommendations expressed in this material are those of responsibility of the author(s) and do not necessarily reflect FAPESP’s view.

\begin{appendices}

\vspace{1.5cm}

\section{Disagreement Metrics}\label{apd:disagreement_metrics}

\citet{krishna2022disagreement} has proposed six metrics to measure disagreement between two explanation methods. However, two of these metrics require the end user to provide a subset of features of interest. We have used only the first four metrics to avoid relying on domain knowledge in selecting these features for our analysis. The following are the four comparison metrics we utilized:

\begin{align}
FA(g_1(x),g_2(x),k) & = \frac{|\, top_k(g_1(x))\,\cap\, top_k(g_2(x))\,|}{k}\label{eq:fa}
\end{align}

\begin{align}
    SA(g_1(x),g_2(x),k) = & \frac{\splitfrac{|\, \bigcup_{s \in S} \{ s \, | \, s \in top_k(g_1(x)) \, \wedge \, s \in top_k(g_2(x)) \,}{\wedge \, sign_k(g_1(x)) = sign_k(g_2(x))  \} \,|}}{k}\label{er:sa}
\end{align}

\begin{align}
    RA(g_1(x),g_2(x),k) = & \frac{\splitfrac{|\, \bigcup_{s \in S} \{ s \, | \, s \in top_k(g_1(x)) \, \wedge \, s \in top_k(g_2(x)) \,}{\wedge \, rank_k(g_1(x)) = rank_k(g_2(x))  \} \,|}}{k}\label{er:ra}
\end{align}

\begin{align}
    SRA(g_1(x),g_2(x),k) = & \frac{\splitfrac{|\, \bigcup_{s \in S} \{ s \, | \, s \in top_k(g_1(x)) \, \wedge \, s \in top_k(g_2(x))}{\wedge \, sign_k(g_1(x) = sing_k(g_2(x) \, \wedge \, rank_k(g_1(x) = rank_k(g_2(x) \} \,|}}{k}\label{er:sra}
\end{align}

Where $top_k(g_i(x))$ represents the $k$ most important features for the prediction of instance $x$ given by the explanation method $g_i$; $rank_k$ returns the top-$k$ most important features ordered according to their relevance; and $sign_k$ the top-$k$ most important features with sign (positive or negative).

\section{Explanation Methods}\label{apd:explanationmethods}

Our study employed nine state-of-the-art feature attribution methods, six of which are gradient-based methods that involve computing the gradient of the output with respect to the input to measure the importance of input features. These gradient-based methods differ in the way they compute the gradients. The six gradient-based methods used in this study are DeepLift~\citep{deeplift}, Guided Back-propagation~\citep{guidedback}, Input X Gradient~\citep{inputxgrad}, Integrated Gradients~\citep{integratedgrad}, Smooth Gradient~\citep{smilkov2017smoothgrad}, and Vanilla Gradients~\citep{vanillagrad}. \citet{MOHAMED2022102239} has detailed these methods, discussing their differences, reliability, and applications.

The Occlusion technique applies a perturbation-based approach to calculate feature attribution~\citep{occlusion}. This type of method works by changing the input and observing the corresponding changes in model prediction to determine the features that significantly impact the model's prediction. 

The Local Interpretable Model-agnostic Explanations (LIME) works by generating interpretable explanations by approximating complex models locally with simpler ones, such as linear models~\citep{lime}. It perturbs the input data around the instance of interest and observes the resulting changes in the model's predictions. LIME then uses these perturbed instances to train a local interpretable model.

KernelSHAP~\citep{kernelshap} operates by computing Shapley values, which represent the marginal contribution of each feature to the difference between the model's prediction and a baseline prediction. The KernelSHAP approximates these Shapley values by employing a kernel-based algorithm that samples subsets of features and computes their contribution to the model's prediction.

\section{Datasets}\label{apd:datasets}

In our experiments, we utilized two real-world datasets.

\citet{Amrieh_2015} dataset originally contained 480 instances and 17 attributes, with 16 attributes used for prediction and one as the target attribute. The predictive attributes predominantly consist of categorical features, including gender, NationalITy, PlaceofBirth, StageID, GradeID, SectionID, Topic, Semester, Relation, raised hands, VisITedResources, AnnouncementsView, Discussion, ParentAnsweringSurvey, ParentschoolSatisfaction, and StudentAbsenceDays. We removed the attributes Topic, NationalITy, PlaceofBirth, SectionID, and GradeID, while the remaining attributes underwent the One-Hot encoding process, removing the first category of each variable. Initially structured as a multiclass variable with three labels (low, medium, and high), the target attribute was adjusted. Instances belonging to the 'medium' class were removed, and the 'high' class was designated as the positive class, with 'low' as the negative class. Consequently, post-preprocessing, we obtained a balanced dataset featuring 12 predictive attributes, a binary target class, and 269 instances.

The Introduction to Programming course dataset was gathered from four classes comprising computer science and computer engineering students. This dataset contains 17 attributes, with 16 predictive attributes and one target attribute. The predictive attributes contain various characteristics, including the grades attained by students during the university admission selection process, demographic information, and details regarding the resolution of programming exercises within the course's system up to the midpoint of the academic semester. To be admitted to the university, students must pass an assessment covering five areas: Writing (essay), Human Sciences, Natural Sciences, Languages and Codes, and Mathematics. For more details on the attributes refer to Table~\ref{tab:dataset2}. Features are standardized by removing the mean and scaling to unit variance for both datasets.

\begin{table}[h]
\caption{Description of the attributes of the introduction to programming dataset.}\label{tab:dataset2}%
\begin{tabular}{@{}lp{6.5cm}@{}}
\toprule
Attribute & Description\\
\midrule
Age    & Student Age.\\
School Type    & This feature indicates whether the student attended a public or private school.\\
Number of people in the family    & Number of family members living in the same house as the student. \\
Program & Computer science or computer engineering\\
Semester & Semester in which the student is taking the introductory programming course.\\
First Grade & Student grade in the first assessment of the semester.\\
Number of Submissions & The number of times students submit their homework through the system provided by the Professor.\\
Number of exercises attempted & Number of exercises the student attempted to answer.\\
Number of exercises completed successfully & Number of exercises that the student answered correctly.\\
Scoring & The system automatically generates a score for the student by evaluating their performance in solving exercises.\\
City & This feature indicates whether the student lived in a small or large city prior to university enrollment.\\
Writing (essay) grade & This feature displays a student's Writing grade in the university admission process.\\
Human Sciences grade & This feature displays a student's Human Sciences grade in the university admission process.\\
Natural Sciences grade & This feature displays a student's Natural Sciences grade in the university admission process.\\
Languages and Codes grade & This feature displays a student's Languages and Codes grade in the university admission process.\\
Mathematics grade & This feature displays a student's Mathematics grade in the university admission process.\\
\botrule
\end{tabular}
\end{table}

\section{Data Distribution of the (dis)agreement score}\label{apd:boxplot}

\begin{figure}[htbp]
\centering
\begin{subfigure}[b]{1\textwidth}
         \centering
         \includegraphics[width=\textwidth]{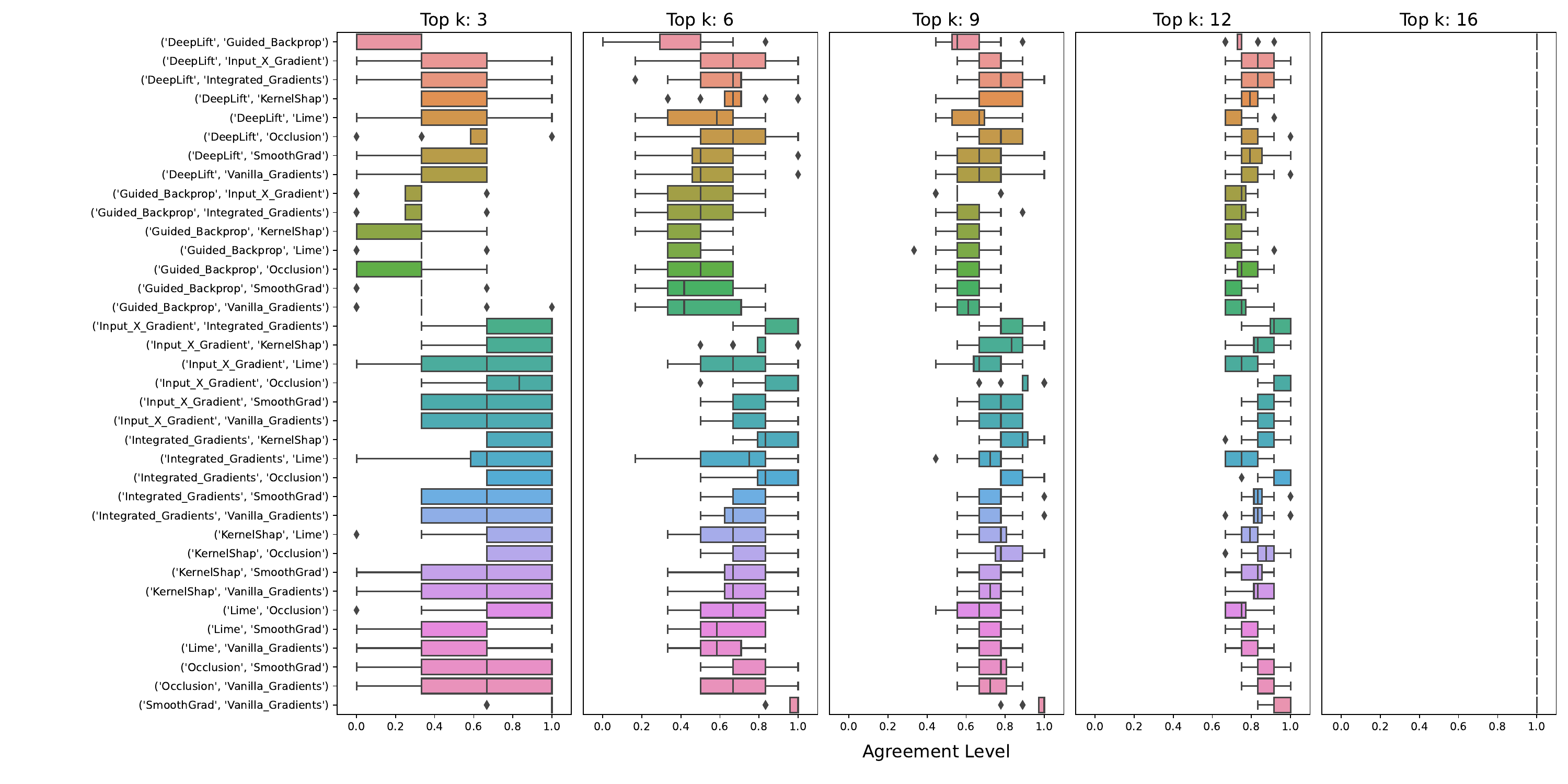}
         \caption{Results for the best model trained with the Introduction to Programming course dataset.}
         \label{fig:fa_boxplot_A}
\end{subfigure}
\qquad
\begin{subfigure}[b]{1\textwidth}
         \centering
         \includegraphics[width=\textwidth]{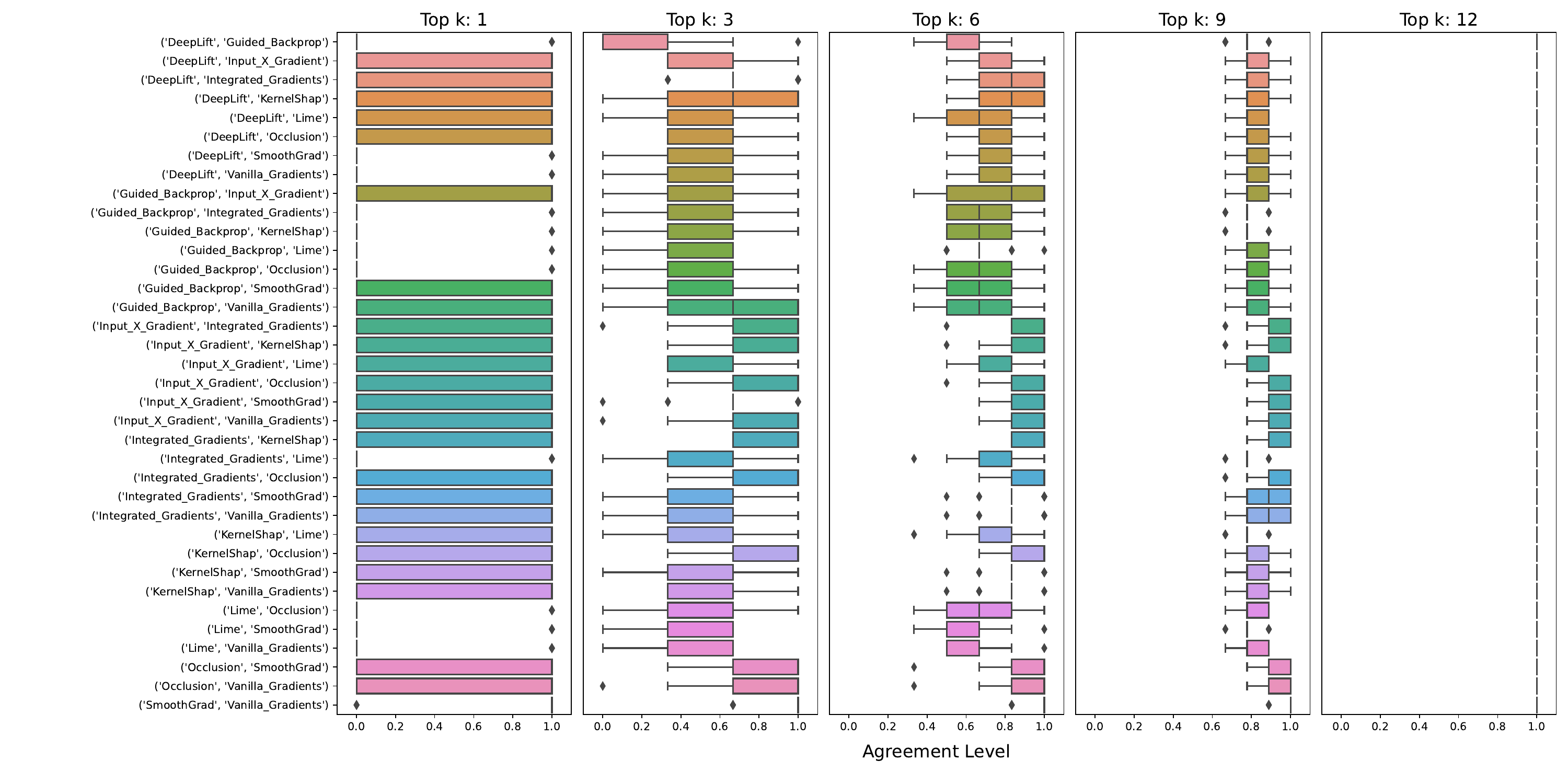}
         \caption{Results for the best model trained with the~\citet{Amrieh_2015} dataset.}
         \label{fig:fa_boxplot_B}
\end{subfigure}     
\caption{Boxplots illustrating the distribution of the disagreement level score by FA metric.}\label{fig:fa_boxplot}
\end{figure}

Figure~\ref{fig:fa_boxplot} visually shows the data distribution of the disagreement score between the pairs of methods for the FA metric and different values to the top-$k$. Figure~\ref{fig:fa_boxplot_A} presents the boxplots for epoch 55 of the model trained with the Introduction to Programming course dataset. This epoch had the best AUC value of 0.85. Furthermore, Figure~\ref{fig:fa_boxplot_B} shows the boxplots for epoch 13 of the model trained with the~\citet{Amrieh_2015} dataset. This epoch had the best AUC value of 1.0.

\section{Example of (dis)agreement between the pairs of methods}\label{apd:heatmap}

\begin{figure}[htbp]
\centering
\begin{subfigure}[b]{0.5\textwidth}
         \centering
         \includegraphics[width=\textwidth]{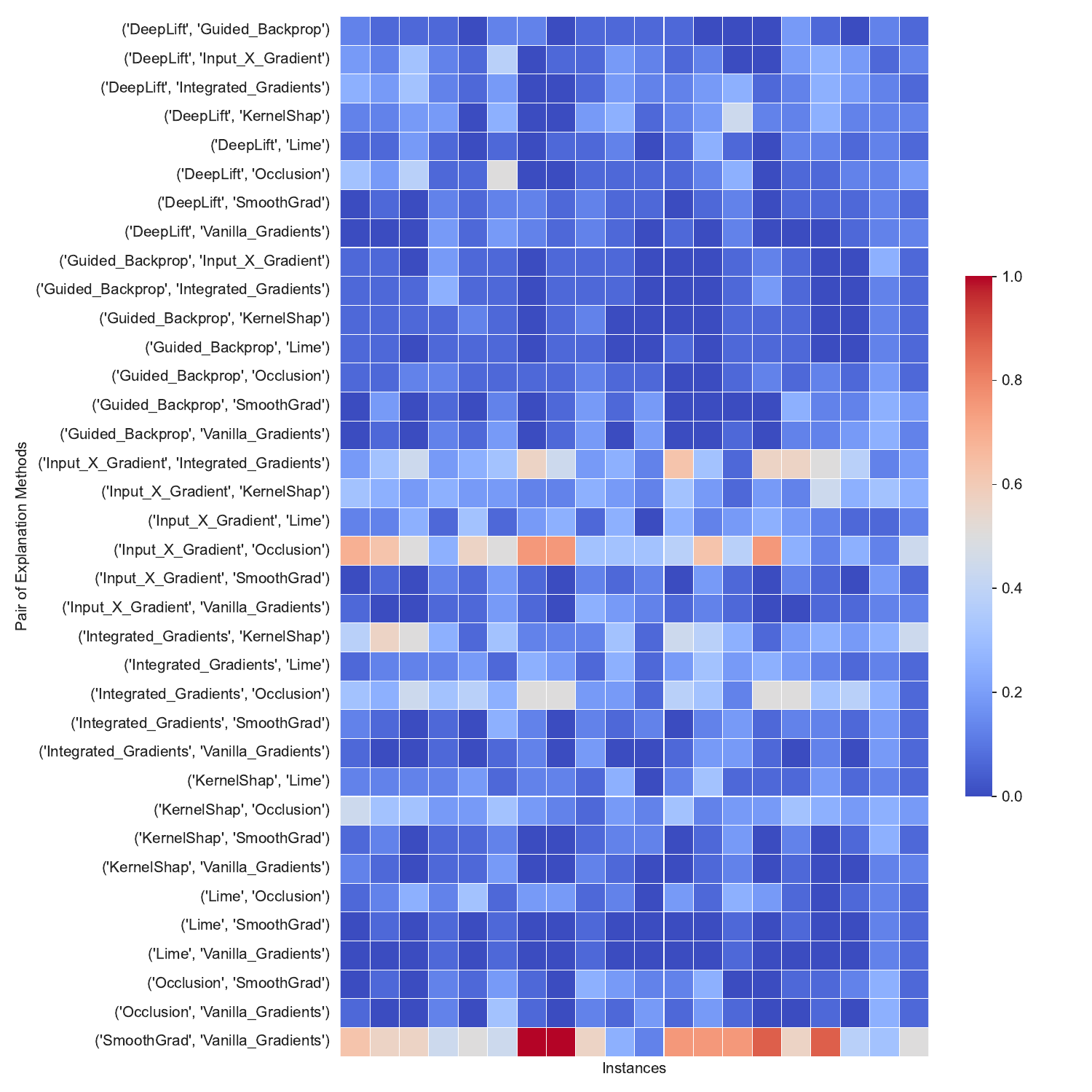}
         \caption{Introd. to Programming course's dataset.}
         \label{fig:heatmap-a}
\end{subfigure}
\qquad
\begin{subfigure}[b]{0.5\textwidth}
         \centering
         \includegraphics[width=\textwidth]{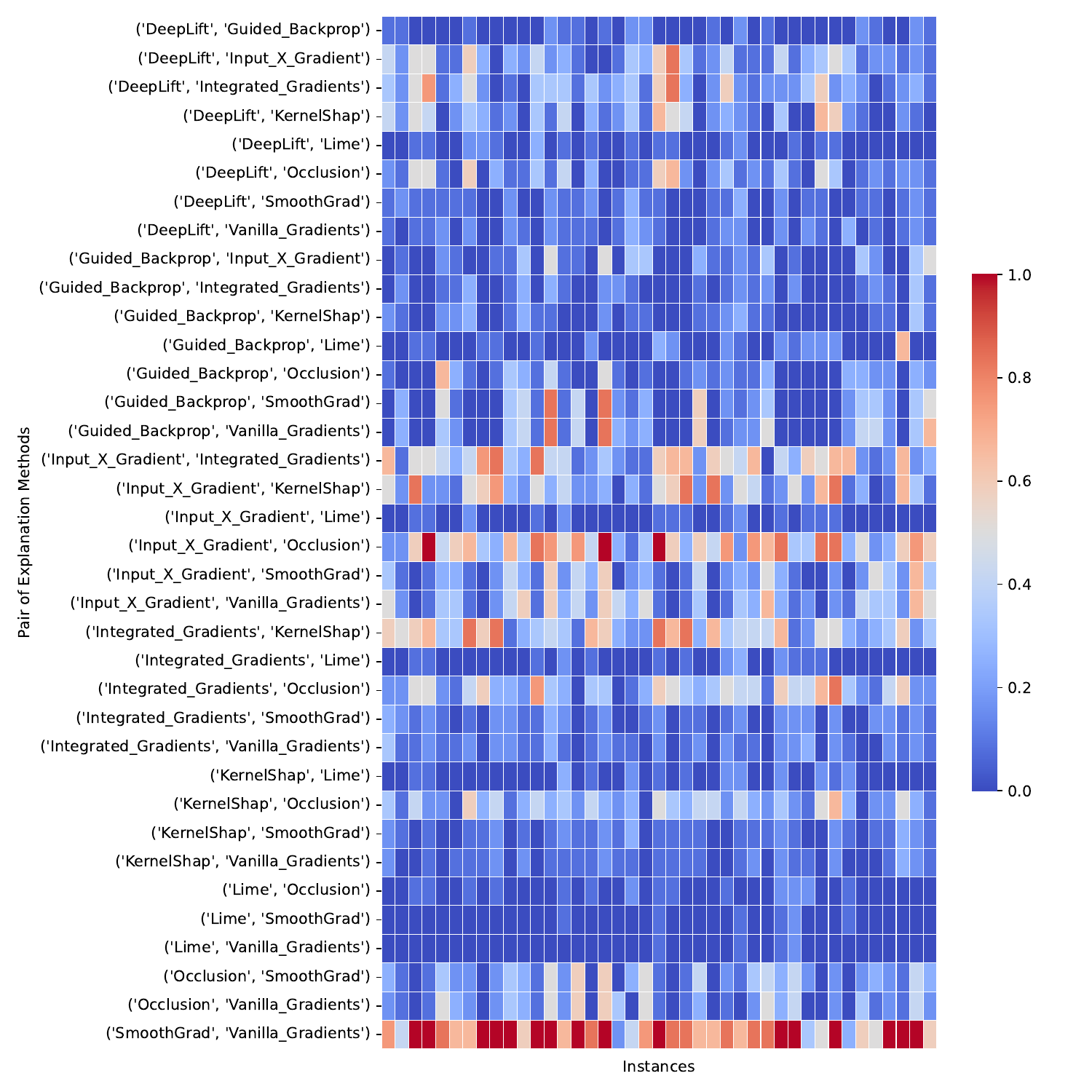}
         \caption{\citet{Amrieh_2015}'s dataset.}
         \label{fig:heatmap-b}
\end{subfigure}     
\caption{Heatmap illustrating the (dis)agreement levels between explanation methods.}\label{fig:heatmap}
\end{figure}

Figure~\ref{fig:heatmap-a} shows a heatmap with the level of (dis)agreement between the pair of methods for epoch 75 of models trained using the Introduction to Programming course's dataset using the metric SRA and $k$ equals 16. Figure~\ref{fig:heatmap-b} shows a heatmap with the level of (dis)agreement between the pair of methods for epoch 13 of models trained using the~\citet{Amrieh_2015}'s dataset using the metric SRA and $k$ equals 12.

\section{AUC vs Disagreement Level}\label{apd:corr}

\begin{figure}[h!]
\centering
\includegraphics[width=0.44\textwidth]{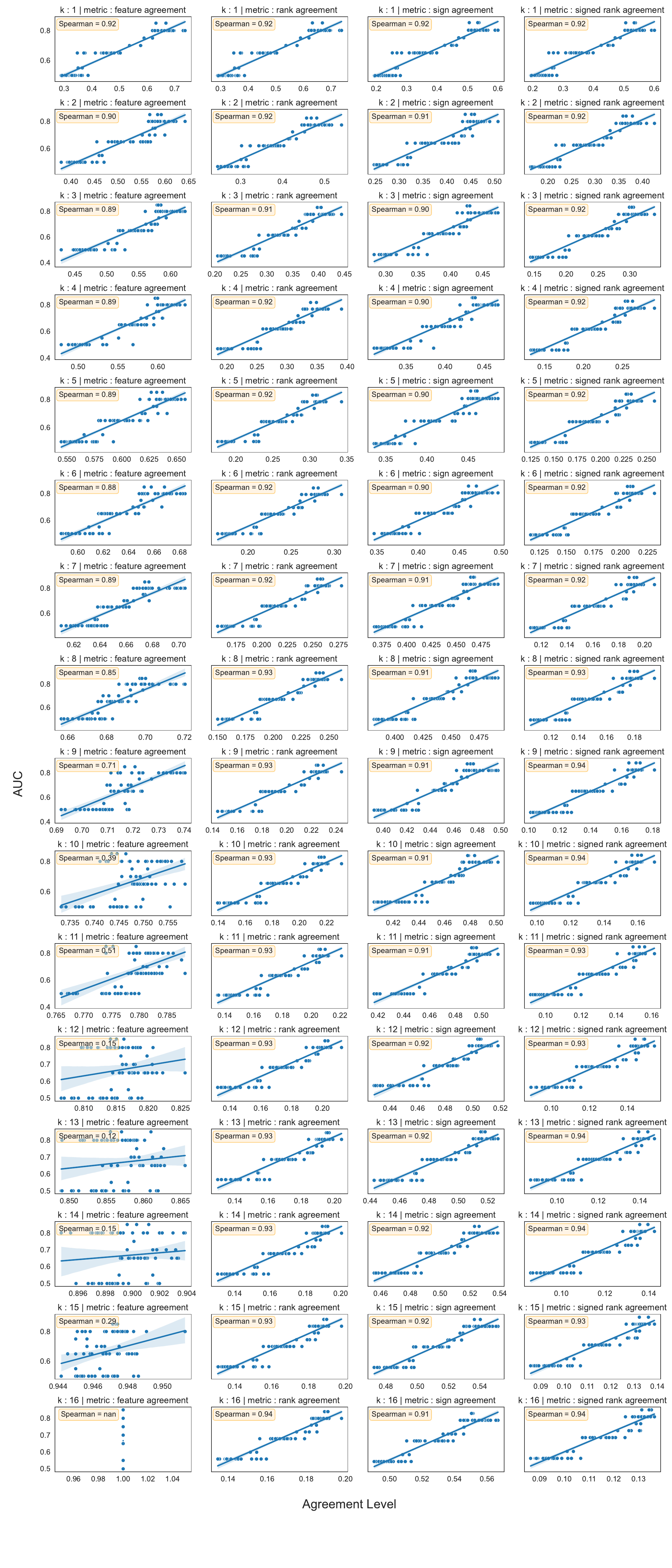}
\caption{Correlation between Model Performance (AUC) and Disagreement Metrics for Models Trained on the Introductory Programming Course Dataset.}\label{fig:corr_cs101_complete}
\end{figure}

Figure~\ref{fig:corr_cs101_complete} shows the correlation between the model performance (AUC) and the (dis)agreement metrics for models trained on the Introductory Programming Course Dataset. The x-axis represents the level of agreement, while the y-axis depicts the model's performance. The charts are organized by disagreement metrics (FA, SA, RA, and SRA), with rows varying the $k$ value from 1 to 16.

\begin{figure}[h!]
\centering
\includegraphics[width=0.44\textwidth]{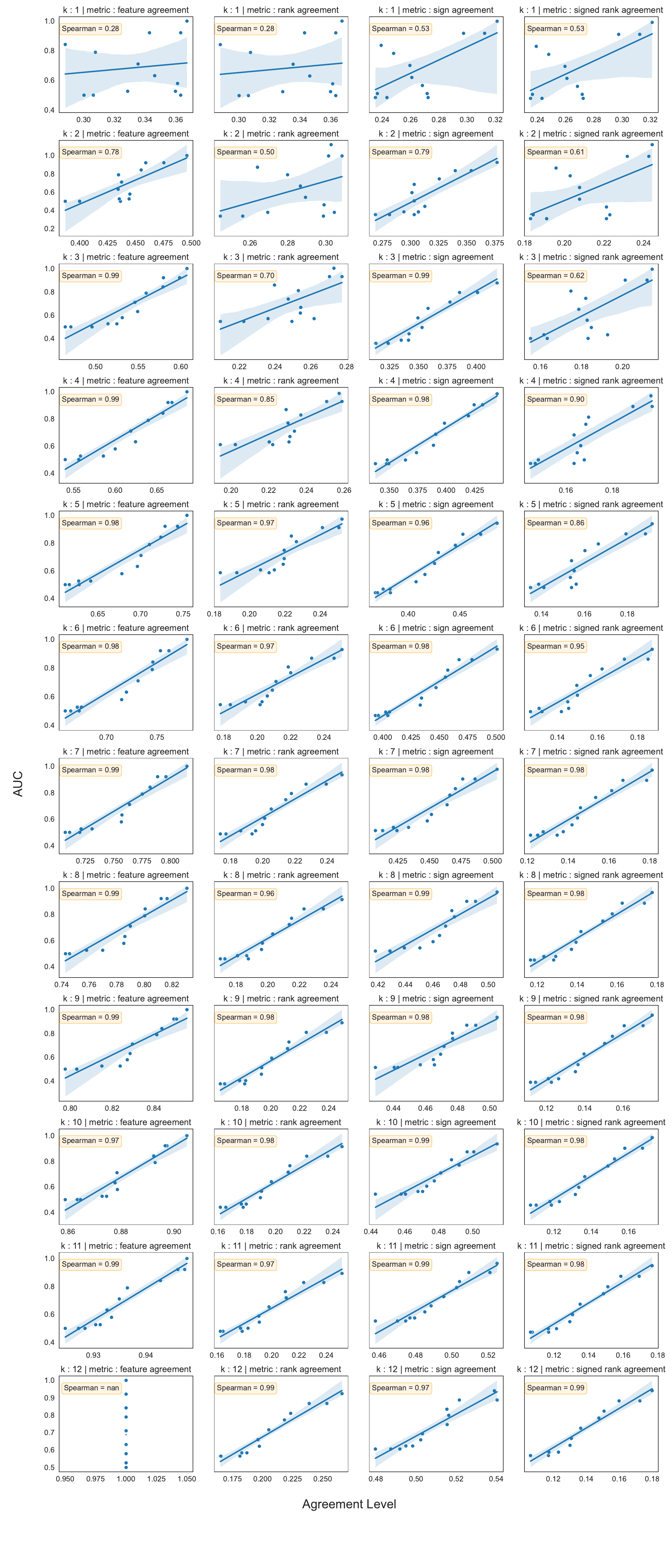}
\caption{Correlation between Model Performance (AUC) and Disagreement Metrics for Models Trained on the~\citet{Amrieh_2015}'s Dataset.}\label{fig:corr_xapi_complete}
\end{figure}

Figure~\ref{fig:corr_xapi_complete} shows the study results for~\citet{Amrieh_2015}'s dataset. The figure illustrates the correlation between model performance (AUC) and the level of agreement across different $k$ values (ranging from 1 to 12, with 12 being the total number of features in the dataset).

\end{appendices}

\bibliography{main}

\end{document}